\documentclass[letterpaper, 10 pt, conference]{styles/ieeeconf}
\IEEEoverridecommandlockouts                             
\usepackage[utf8]{inputenc}
\usepackage[T1]{fontenc}
\usepackage[english]{babel}

\usepackage{xspace}

\usepackage{url}
\usepackage{color}
\usepackage{wrapfig}
\usepackage{placeins}
\newif\iffigs
\figstrue %

\usepackage[fleqn]{amsmath}
\usepackage{float}
\usepackage{setspace}
\usepackage{graphicx}
\usepackage{mathrsfs}
\usepackage{amssymb}
\usepackage{nicefrac}
\usepackage{algorithm}
\usepackage[noend]{algpseudocode}

\makeatletter
\newcommand\fs@spaceruled{\def\@fs@cfont{\bfseries}\let\@fs@capt\floatc@ruled
  \def\@fs@pre{\vspace{0.4\baselineskip}\hrule height.8pt depth0pt \kern2pt}%
  \def\@fs@post{\vspace{-0.4\baselineskip}\kern2pt\hrule\relax\vspace{-12pt}}%
  \def\@fs@mid{\kern2pt\hrule\kern2pt}%
  \let\@fs@iftopcapt\iftrue}
\makeatother

\usepackage{tabularx}
\usepackage{multirow}
\usepackage{booktabs}
\usepackage{colortbl}
\usepackage{siunitx}

\usepackage{etoolbox}
\makeatletter
\patchcmd{\@makecaption}
{\scshape}
{}
{}
{}
\makeatother

\let\labelindent\relax
\usepackage[inline]{enumitem}

\usepackage{listings}
\lstnewenvironment{itemlisting}[1][]
 {%
  \mbox{}
  \vspace*{-\baselineskip}
  \lstset{
    xleftmargin=\leftmargin,
    linewidth=\linewidth,
    #1
  }%
 }
 {}
\usepackage{multicol}
\usepackage{footmisc}

\usepackage{lipsum}
\usepackage{xcolor}
\usepackage{subcaption}
\usepackage{csquotes}
\usepackage[
	maxbibnames=99,
	maxcitenames=2,
	natbib=true,
	style=numeric-comp,
	backend=biber,
	sorting=none,
	giveninits=true,
	url=false,
	doi=false,
	eprint=false,
	isbn=false,
]{biblatex}

\usepackage{xurl}

\usepackage[pdfa,colorlinks,bookmarksopen,bookmarksnumbered,allcolors=black,urlcolor=blue]{hyperref}

\usepackage[nameinlink,capitalise]{cleveref}
\crefname{line}{line}{lines}
\crefname{figure}{Fig.}{Figs.}
\Crefname{figure}{Fig.}{Figs.}
\crefname{equation}{Eq.}{Eqs.}
\Crefname{equation}{Eq.}{Eqs.}
\crefname{section}{Sec.}{Secs.}
\Crefname{section}{Sec.}{Secs.}
\crefname{definition}{Def.}{Defs.}
\Crefname{definition}{Def.}{Defs.}
\crefname{algorithm}{Alg.}{Algs.}
\Crefname{algorithm}{Alg.}{Algs.}
\crefname{assumption}{Asm.}{Asms.}
\Crefname{assumption}{Asm.}{Asms.}
\crefname{subassumption}{Asm.}{Asms.}
\Crefname{subassumption}{Asm.}{Asms.}
\Crefname{problem}{Problem}{Problems}
\crefname{problem}{Problem}{Problems}

\definecolor{pRRTC_color_threads}{HTML}{328535}%
\newcommand{\pRRTCcolornamethreads}{green}
\definecolor{pRRTC_color_groups}{HTML}{ff77b4} 
\newcommand{\pRRTCcolornamegroups}{pink}

\DeclareMathOperator*{\argmin}{argmin}
\DeclareMathOperator*{\argmax}{argmax}

\usepackage{flushend}

\newif\ifanon
\anonfalse

\ifanon
\else
\fi

\title{\LARGE \bf pRRTC: GPU-Parallel RRT-Connect for Fast,\\ Consistent, and Low-Cost Motion Planning}

\ifanon
\else
\author{Chih H. Huang$^{1*}$, Pranav Jadhav$^{2*}$, Brian Plancher$^{3,4}$, and Zachary Kingston$^{2}$%
\thanks{This material is based upon work supported by the National Science Foundation (under Award 2411369). Any opinions, findings, conclusions, or recommendations expressed in this material are those of the authors and do not necessarily reflect those of the funding organizations.}%
\thanks{$^{1}$CHH is with Columbia College, Columbia University, New York, NY. {\tt\footnotesize chih.h@columbia.edu}}%
\thanks{$^{2}$PJ and ZK are with the Department of Computer Science, Purdue University, IN. {\tt\footnotesize \{jadhav14, zkingston\}@purdue.edu}}%
\thanks{$^{3,4}$BP is with Dartmouth College, Hanover, NH and Barnard College, Columbia University, New York, NY. {\tt\footnotesize plancher@dartmouth.edu}}%
\thanks{$^{*}$Equal Contribution.}
}
\fi

\begin{document}
\maketitle
\thispagestyle{empty}
\pagestyle{empty}
\setcounter{footnote}{3}

\begin{abstract}
    Sampling-based motion planning algorithms, like the Rapidly-Exploring Random Tree (RRT) and its widely used variant, RRT-Connect, provide efficient solutions for high-dimensional planning problems faced by real-world robots. However, these methods remain computationally intensive, particularly in complex environments that require many collision checks. To improve performance, recent efforts have explored parallelizing specific components of RRT such as collision checking, or running multiple planners independently. However, little has been done to develop an integrated parallelism approach, co-designed for large-scale parallelism. In this work we present pRRTC, a RRT-Connect based planner co-designed for GPU acceleration across the entire algorithm through parallel expansion and SIMT-optimized collision checking. We evaluate the effectiveness of pRRTC on the MotionBenchMaker dataset using robots with 7, 8, and 14 degrees of freedom (DoF). Compared to the state-of-the-art, pRRTC achieves as much as a 10× speedup on constrained reaching tasks with a 5.4× reduction in standard deviation. pRRTC also achieves a 1.4× reduction in average initial path cost. Finally, we deploy pRRTC on a 14-DoF dual Franka Panda arm setup and demonstrate real-time, collision-free motion planning with dynamic obstacles. We open-source our planner to support the wider community.

\end{abstract}

\section{Introduction} \label{sec:intro}
Motion planning is a fundamental problem in robotics that involves finding collision-free motion paths through a robot’s configuration space~\cite{lavalle2006planning, kavraki2016motion,orthey2023sampling}. While there are many approaches to planning, sampling-based motion planning (SBMP) approaches are widely used due to their generality and efficiency in higher-dimensions. One of the most popular SBMP algorithms is the Rapidly-Exploring Random Tree (RRT)~\cite{lavalle2001rapidly} and its bidirectional variant RRT-Connect~\cite{kuffner2000rrt}. 
RRT-Connect's design biases it towards quickly solving problems involving large open spaces, even in high dimensions. However, its performance suffers in cluttered environments such as in constrained reaching tasks~\cite{orthey2023sampling}.

One potential avenue for improving planning performance is through the use of parallelism. Existing parallelized approaches fall into one of three categories:
(1) \emph{high-level parallelism} by executing multiple RRT instances simultaneously~\cite{otte2013c, hidalgo2018quad}, (2) \emph{mid-level parallelism} by executing multiple sample-and-grow iterations simultaneously~\cite{ichnowski2012parallel}, and (3) \emph{low-level parallelism} over primitive operations such as collision checking~\cite{thomason2024motions, bialkowski2011massively,murray2016robot,murray2016microarchitecture}. While prior works have mostly focused on one form of parallelism, little has been done on integrating an approach over multiple levels of parallelism to maximize performance on modern parallel hardware (e.g., GPUs). 

To address this gap, we introduce pRRTC, a parallel RRT-Connect-based algorithm designed for GPU acceleration across multiple parallelism levels: mid-level planning iteration parallelism and low-level primitive operation parallelism. Our approach's performance is driven by three key design decisions:
(1) concurrent sampling, expansion and connection of start and goal trees via GPU threads, (2) SIMT-optimized collision checking to quickly validate edges, inspired by the SIMD-optimized validation of~\citet{thomason2024motions}, and (3) efficient memory management between block and thread level parallelism, ultimately reducing expensive memory transfer overheads.
Compared to other GPU-based SBMP algorithms (e.g.,~\cite{bialkowski2011massively,ichter2017group,hidalgo2018quad}), our approach supports high-dimensional manipulators and achieves sub-millisecond planning performance.

\begin{figure}[!t]
    \centering
    \includegraphics[width=0.95\linewidth]{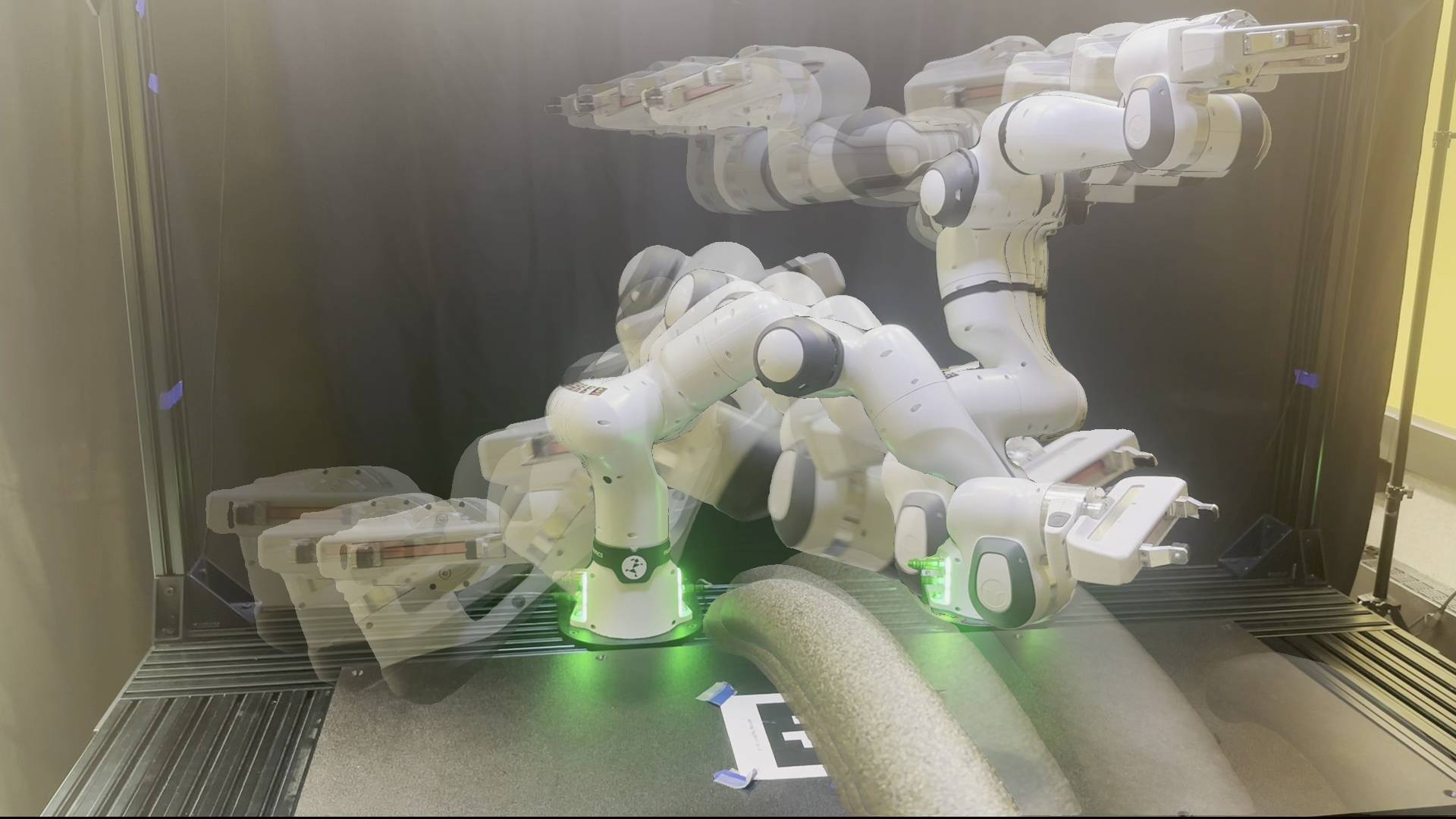}
    \caption[]{pRRTC achieves real-time replanning on a 14-DoF dual-arm Franka Panda setup, enabling dynamic obstacle avoidance. Shown here is a superimposed image from several frames of our supplementary video, illustrating the robot arms successfully avoiding moving pool noodles.}
    \label{fig:hardware}
    \vspace{-15pt}
\end{figure}

By integrating parallelism across planning iterations and primitive operations, pRRTC achieves fast, consistent, and efficient planning. %
We evaluate pRRTC against state-of-the-art CPU- and GPU-based motion planners on the MotionBenchMaker dataset~\cite{chamzas2021motionbenchmaker} using robots with 7, 8, and 14 degrees of freedom (DoF). pRRTC achieves as much as a 10× speedup on constrained reaching tasks over state-of-the-art planners in complex environments. 
Importantly, pRRTC exhibits a 5.4× reduction in planning time standard deviation across all problems, indicating consistency across a variety of environments and robots.
pRRTC also produces low-cost initial paths, achieving a 1.4× reduction in average initial path cost compared to existing approaches. These results highlight the benefits of software-hardware co-designed GPU parallelism for accelerating SBMPs. 
Finally, we deploy pRRTC on a 14 DoF dual Franka Panda arm setup and demonstrate real-time, collision-free motion planning with dynamic obstacles (see~\cref{fig:hardware}). We open-source our planner at: 
\ifanon
\url{https://anonymous.4open.science/r/pRRTC-ICRA-26}
\else
\url{https://github.com/CoMMALab/pRRTC}
\fi%
.

\section{Background and Related Work} \label{sec:related}
\subsection{The RRT-Connect Planning Algorithm}

There is a vast array of SBMP approaches each with their own modifications to core operations of sampling, nearest neighbors search, steering, and collision checking ~\cite{Choset2005,lavalle2006planning,orthey2023sampling}. In this work, we focus on the RRT and RRT-Connect algorithms due to their fast single-query performance.

Many modifications have been proposed to the RRT algorithm, such as the growth of multiple trees~\cite{kuffner2000rrt,wang2010triple,wang2010adaptive}, biased sampling~\cite{yershova2005dynamic,urmson2003approaches,liu2019partition,rodriguez2006obstacle}, and environment preprocessing~\cite{shu2019locally}.
RRT-Connect~\cite{kuffner2000rrt} is a widely used variant that simultaneously expands two trees: one rooted at the start configuration and the other at the goal configuration.
The balanced variant of RRT-Connect~\cite{kuffner2005efficient} selects the smaller tree for tree growth at each iteration, which is more efficient for problems with unbalanced difficulty around the start and goal configurations.
Each iteration of the algorithm also includes an additional connect operation, where the newly added node is greedily extended toward the opposing tree until a collision is detected.
The planner terminates when the two trees successfully connect, forming a complete path.
Additionally, sampling heuristics such as dynamic domain sampling~\cite{yershova2005dynamic} can be easily integrated to improve performance.

RRT-Connect is well known for accelerating planning in large, open spaces~\cite{kuffner2000rrt}. However, many problems contain \emph{narrow passages} created by environment geometry (e.g., when a robot must reach into a container), where motion is heavily constrained. In such cases, a large number of samples must be evaluated, motivating the extension of trees in parallel. A critical component of this process is motion validation, which depends on collision checking.

Collision checking is empirically recognized as a primary bottleneck in motion planning~\cite{bialkowski2011massively}. This has motivated diverse algorithmic strategies, including hierarchical collision detection~\cite{li1998incremental}, restructured computations for faster expected performance~\cite{Sanchez2003}, delayed or lazy checking~\cite{sanchez2002delaying, hauser2015lazy}, exact and swept-volume approaches~\cite{Schwarzer2004, joho2024neural, son2025neuralsvcd}, and caching techniques~\cite{bialkowski2013efficient}. Given the independence of collision-checking operations across candidate samples, parallelism offers a natural and promising direction for accelerating this stage of the planning pipeline.

\subsection{Parallel Acceleration of Planning Algorithms}

Parallelization of planning algorithms has been studied since the advent of the field~\cite{Barraquand1990,Henrich1997}.
In particular, Amato et al.~\cite{amato1999probabilistic} noted early on that SBMPs (e.g.,~\cite{Kavraki1996}) were ``embarrassingly parallel.'' We categorize the parallelism of SBMPs into three categories: high-, medium-, and low-level parallelism.
High-level parallelization refers to running multiple isolated planners in parallel; medium-level parallelization refers to running multiple iterations within the same planning framework in parallel; low-level parallelization refers to parallelizing primitive operations within each iteration of a single planning algorithm.

In terms of CPU parallelism, most parallel planners achieve high- or medium-level parallelization, such as running many instances of the planner simultaneously~\cite{otte2013c}, running RRT iterations in parallel on one tree~\cite{ichnowski2012parallel, Xiao2017}, and parallelizing search with forests of trees~\cite{Plaku2005, jacobs_scalable_method_2012}.
With regard to low-level parallelism, VAMP~\cite{thomason2024motions,ramsey2024} utilizes vectorized motion validation through CPU Single Instruction, Multiple Data (SIMD) instructions leading to data parallelism in the collision checking process.
However, CPU-based parallelism is limited in scope due to a relatively small number of cores and the limitations of current CPU SIMD operations.

GPUs provide massive parallelism through the Single Instruction, Multiple Thread (SIMT) model over thousands of cores.
While powerful, GPUs require careful algorithm design to fully leverage the hardware.
High-level parallelization strategies have been proposed by~\citet{hidalgo2018quad} and~\citet{jacobs2013scalable}, offering substantial speedups in environments with narrow passages.
A number of different GPU-based SBMPs have been proposed such as GMT~\cite{ichter2017group}, Pk-RRT~\cite{tran2023gpu}, Kino-PAX~\cite{perrault2024kino}, and the RRT-like planner used by cuRobo~\cite{sundaralingam2023curobo} which leverages medium-level parallelism for parallel sampling, rollouts, and tree growth.
Work on low-level parallelism has been focused on accelerating collision checking through clustering~\cite{pan2012gpu}, spatial hashing~\cite{pabst2010fast}, hierarchy construction~\cite{lauterbach2010gproximity}, and checking discretized configurations along a motion in parallel~\cite{bialkowski2011massively,murray2016robot,murray2016microarchitecture}.

Finally, we note that outside of SBMP algorithms there have similarly been many developments on GPU-parallelization of other classes of planning and control algorithms including both search-based planning~\cite{zhou_massively_parallel,fukunaga_survey_parallel_2017,fukunaga_parallel_statespace_2018}, as well as optimal control-based techniques~\cite{williams2017model,PlancherParallelDDP,sundaralingam2023curobo,adabag2024mpcgpu,jeon2024cusadi,chari2024fast}. %

Despite these many advances and clear evidence of performance gains, existing works that parallelize SBMP algorithms only utilize parallelism at one specific level, limiting their overall performance.
To overcome this gap, we developed pRRTC, a GPU-based planner that leverages co-designed parallelism across multiple levels to surpass state-of-the-art performance.

\section{pRRTC Algorithm} 
\label{sec:design}
\iffigs
\begin{figure*}[!t]
   \centering
   \vspace{1em}
   \includegraphics[width=0.99\linewidth]{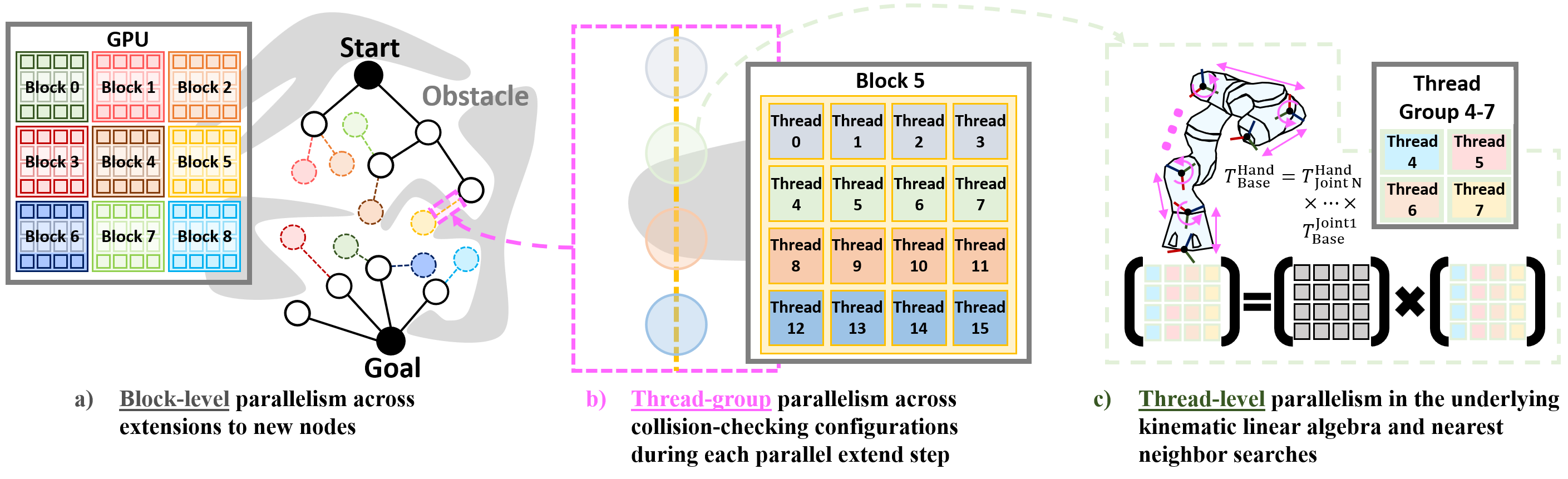}
   \vspace{-5pt}
   \caption{A conceptual illustration of pRRTC in an abstract configuration space operating across three levels of GPU-parallelism. a) Separate GPU blocks extend both the start and goal trees with multiple samples in parallel. b) Within each block, parallel groups of four threads compute collision checks across the multiple configurations that correspond intermediate states of a motion. c) Individual threads parallelize the underlying kinematic linear algebra needed for these collision checks. Individual threads also parallelize nearest neighbors search.}
   \label{fig:design}
   \vspace{-15pt}
\end{figure*}
\fi

In this section, we discuss the algorithmic design and implementation of pRRTC as diagrammed in~\cref{fig:design} and~\cref{alg:prrtc,alg:extend}.
Overall, pRRTC retains similar primitive operations and structure as the vanilla RRT-Connect algorithm, but it implements aggressive mid- and low-level parallelism designed to leverage the computational structure of modern GPUs.
At the mid-level, pRRTC runs hundreds of parallel RRT-Connect iterations asynchronously across both trees (\cref{alg:prrtc} lines 4-19). This enables pRRTC to explore the configuration space faster and find higher quality paths. At the low-level, pRRTC parallelizes nearest neighbors (NN) search and discretized edge collision checking (CC).

This parallelism is realized via blocks of parallel threads on the GPU. Each block is responsible for running independent RRT-Connect iterations with a constant number of threads $t_1,\dots,t_n$ to parallelize low-level operations.
NN search and CC against the robot itself and environment are parallelized across individual threads (shown in \textcolor{pRRTC_color_threads}{\pRRTCcolornamethreads}). For forward kinematics (FK), threads are logically organized into groups of four to parallelize matrix operations (shown in \textcolor{pRRTC_color_groups}{\pRRTCcolornamegroups}) building on previous GPU designs~\cite{sundaralingam2023curobo}.

Our block and thread organization is critical for efficiency. Threads within a block share a physical processor and low-latency shared memory, enabling fast synchronization for fine-grained parallel operations. Since independent RRT-Connect iterations require little coordination, blocks can operate asynchronously and access the global trees only when needed. This allows each block to explore the search space at its own pace with minimal synchronization overhead, while still leveraging the progress made by others. To handle potential race conditions on the shared global trees, pRRTC employs atomic primitives to ensure new configurations are written safely to memory.

\captionsetup[algorithm]{font=normalsize}
\floatstyle{spaceruled}%
\restylefloat{algorithm}%
\begin{algorithm}[!t]
\begin{spacing}{1.25}
\begin{algorithmic}[1]
\caption{{pRRTC($c_{start}, c_{goal}, \delta$)}} 
\label{alg:prrtc}
\State $T_a$.init(), $T_b$.init() 
\State $T_a$.add($c_{start}$), $T_b$.add($c_{goal}$)
\For{\textbf{block index} $i = 0\dots N_{samples}$} \textbf{in parallel}
   \For{iteration $iter \gets 0\dots \text{max\_iters}$}
        \State $T_{s}^i, T_{o}^i$ $\gets$ $\argmin (|T_a|, |T_b|)$, $\argmax (|T_a|, |T_b|)$
        \State $c_{rand}^i$ $\gets$ RANDOM\_CONFIG()
        \State $c_{nns}^i$ $\gets$ \textcolor{pRRTC_color_threads}{NEAREST\_NEIGHBOR($c_{rand}^i$, $T_s^i$)}
        \State $c_{new}^i$, $\text{valid}^i$ $\gets$
        \textcolor{pRRTC_color_groups}{pEXTEND($c_{rand}^i$, $c_{nns}^i$, $\delta$)}
        \If{$\text{valid}^i$} \textcolor{gray}{// Add and Greedily extend to $T_o^i$}
            \State $T_{s}^i$.add($c_{new}^i$)
            \State $c_{nno}^i$ $\gets$ \textcolor{pRRTC_color_threads}{NEAREST\_NEIGHBOR($c_{new}^i$, $T_o^i$)}
            \State $n_{ext} \gets$ DISTANCE($c_{new}^i$, $c_{nno}^i$) / $\delta$
            \State $v_{ext} \gets (c_{nno}^i-c_{new}^i) / \delta$
            \For{$i_{ext} \gets 1 \dots n_{ext}$}
                \State $c_{ext}^i \gets c_{new}^i+v_{ext}$ 
                \State $c_{new}^i, \text{valid}^i \gets$ \textcolor{pRRTC_color_groups}{pEXTEND($c_{new}^i$, $c_{ext}^i$, $\delta$)}
                \If{$\text{valid}^i$} $T_{s}^i$.add($c_{new}^i$)
                \Else \;\textbf{break}
                \EndIf
            \EndFor
            \If{$i_{ext}=n_{ext}$} \Return PATH($T_a$, $T_b$)
            \EndIf 
       \EndIf
   \EndFor
   \State \Return Failed
\EndFor
\end{algorithmic}
\end{spacing}
\end{algorithm}

Finally, when serial operations are required, they are executed by a designated lead thread $t_1$ within each block. This role is primarily reserved for intermittent, higher-level control flow tasks, such as selecting which tree to expand, sampling new states, and updating the global tree. %

The complete algorithm is shown in~\cref{alg:prrtc} and~\cref{alg:extend} where
$c$ refers to joint space configurations, $\delta$ is the extension range parameter, $T_a$ is the start tree, $T_b$ is the goal tree, $T_s^i$ and $T_o^i$ refer to the tree being extended and the opposite tree for block $i$, and $N_{cc}$ is the collision checking resolution. While omitted above for improved readability and clarity of our overall approach, we also use the dynamic domain sampling heuristic described in Yershova et al.~\cite{yershova2005dynamic}.

\floatstyle{spaceruled}%
\restylefloat{algorithm}%
\begin{algorithm}[!t]
\begin{spacing}{1.25}
\begin{algorithmic}[1]
\caption{pEXTEND($c_{rand},c_{nn},\delta$)}
\label{alg:extend}
\State valid $\gets$ True
\State $c_{new}$ $\gets$ 
INTERPOLATE($c_{rand},c_{nn},\delta$)
\For{$i \gets 1 \dots N_{cc}$} \textcolor{pRRTC_color_groups}{in parallel thread-groups}
    \State $c_{check}$ $\gets$ INTERPOLATE($c_{rand},c_{new},i/N_{cc}*\delta$)
    \State valid $\&=$ \textcolor{pRRTC_color_threads}{COLLISION\_CHECK($c_{check}$)}
\EndFor
\State \Return $c_{new}$, valid
\end{algorithmic}
\end{spacing}
\end{algorithm}

\subsection{SIMT Collision Checking}\label{sec:simtcc}
As mentioned previously, collision checking (CC) is widely recognized as a major computational bottleneck in motion planning~\cite{bialkowski2011massively}, particularly when validating long or numerous edges in high-dimensional configuration spaces. Many prior approaches aim to accelerate CC by spatially distributing checks along an edge using binary subdivision~\cite{sanchez2002using} and SIMD parallelism~\cite{thomason2024motions}. We build on these approaches and use GPU parallelism to check a set of discrete points along an edge simultaneously.

Each edge in our planner is discretized according to a pre-defined CC resolution, and the entire set of points along that edge is validated in parallel. Specifically, we assign a group of four threads to each configuration to exploit the 4×4 matrix operations required in forward kinematics (FK) inspired by past GPU designs~\cite{sundaralingam2023curobo}. For example, if an edge is discretized into 32 configurations, the GPU block will contain $32 \times 4=128$ threads. Since the maximum extension length is bounded by the planner's RRT range parameter, we can preallocate these threads at kernel launch time, avoiding iteration and synchronization overhead. Furthermore, unlike approaches that store all intermediate matrices, we minimize shared memory usage by overwriting intermediate results whenever possible. For high-DOF robots such as Baxter, this strategy reduces shared memory consumption by up to 7×, enabling a larger number of concurrent blocks per streaming multiprocessor (SM), critical to overall planning throughput.

We further accelerate this routine by enabling early exits. When a collision is detected along an edge, the candidate edge is invalidated and all checks along that edge are halted. While conceptually simple, implementing this efficiently on the GPU is challenging as synchronizing across all threads in a block introduces undesirable overheads. To avoid this, we leverage warp\footnote{A ``warp'' represents 32 contiguous threads on the same GPU-core. These threads work in lock-step due to the design of NVIDIA GPU hardware and as such have native implicit synchronization at the hardware level.} primitives, which allow fast coordination among groups of 32 threads through low-latency GPU memory. Since our implementation assigns four threads per CC, each warp can evaluate and efficiently enable early exit synchronization for eight checks in parallel. For environment collisions, where the workload is balanced, threads assume all peers remain active. For self-collisions, where workloads vary due to differing numbers of collision spheres, threads use a single-cycle \textit{active mask}~\cite{NVIDIA25CUDA} to identify which threads are active and share results only among them.

We use the \texttt{foam} tool~\cite{coumar2025foam} to compute a set of bounding spheres representing the robot offline. These spheres are used at runtime in a two-stage collision checking process. First, a low-resolution FK and CC pass is run, flagging links that may be in collision. Only those links are re-checked using a high-resolution pass, which leverages finer-grained sphere models for more precise analysis. This hierarchical approach reduces expensive CC calls while preserving accuracy.

For static environments, such as those from MotionBenchMaker~\cite{chamzas2021motionbenchmaker}, we represent obstacles using primitive geometry. In these settings, the small number of obstacles allows the full set of environment primitives to be cached in low- or mid-level GPU memory throughout most of the planning process, enabling low-latency access across threads in each block. For real-world experiments, we integrate with Nvblox~\cite{millane2024nvblox} to perform CC against dynamically updated Euclidean Signed Distance Field (ESDF) maps.

\subsection{Nearest-Neighbor (NN) Search }
\label{sec:nn}
We use divide-and-conquer parallelism to accelerate our NN search. Within a block of $n$ threads, pRRTC divides each NN query into $n$ subproblems, with each thread assigned to search a disjoint subset of the existing tree for a local NN. Once these are found, pRRTC constructs the final global NN using a tree-based parallel reduction~\cite{harris2007optimizing}. For a given tree of size $T$, serial NN requires $T$ comparisons while parallel NN requires $\lceil \nicefrac{T}{n} + \log (\nicefrac{n}{2}) \rceil$.
This approach avoids significant computational overhead and remains competitive with serial approaches even with low total node counts.

\section{Experiments} \label{sec:experiments}
\subsection{Methodology}
We evaluate pRRTC against state-of-the-art CPU and GPU baselines. On the CPU we use two CPU-based RRT-Connect implementations: the RRT-Connect provided by the Open Motion Planning Library (OMPL-RRTC)~\cite{sucan2012the-open-motion-planning-library} and the implementation provided by VAMP (VAMP-RRTC)~\cite{thomason2024motions}.
To ensure a fair evaluation, OMPL-RRTC was built with VAMP as the motion validation backend, as VAMP has shown orders of magnitude speedup compared to the default setup of OMPL. 
We choose these implementations because they are known to be the best performing planners on the MotionBenchMaker dataset in terms of planning time and solution cost~\cite{orthey2023sampling,thomason2024motions,wilson2024nearest}.
We also compare against two state-of-the-art GPU-based planners: Curobo~\cite{sundaralingam2023curobo} and Global Tensor Motion Planning (GTMP)~\cite{le2024global} with straight-line interpolation (GTMP Straight) and Akima splines (GTMP Akima). 
All experiments were conducted on an x86-based desktop computer with an AMD Ryzen Threadripper PRO 5965WX 24-Core CPU and an NVIDIA GeForce RTX 4090 GPU. We leverage all parallelism available in these planners whether on the CPU or GPU. Hyperparameters are controlled across planners to ensure equivalent implementations with identical collision checking resolution and motion extension range. Planner-specific hyperparameters are chosen to maximize percentage of problems solved and to minimize planning time. We measure the planning time of all algorithms excluding the environmental setup step for both the CPU and GPU to maintain a fair and consistent comparison. We use identical multi-dimensional Halton sequences for configuration sampling across pRRTC, VAMP-RRTC, and OMPL-RRTC, and default samplers for Curobo and GTMP.

\begin{figure}[t]
    \centering
    \vspace{5pt}
    \includegraphics[width=0.95\linewidth]{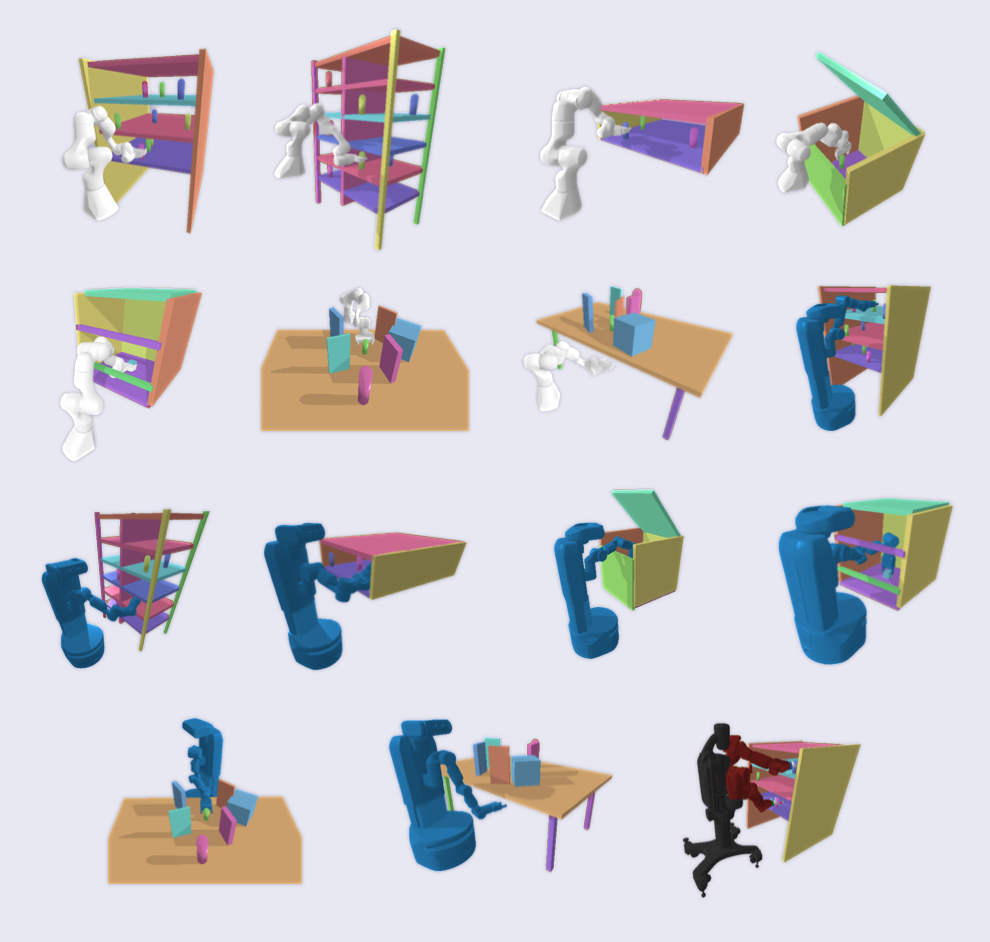}
    \caption[]{The MotionBenchMaker scenes provide diverse manipulation problems for 7, 8, and 14 DoF robots. The problems include counter top manipulation, accessing shelves, and constrained reaching. Panda and Fetch have 7 scenes with 100 problems each, while Baxter has 3 scenes with 500 problems each.}
    \label{fig:motionBenchMakerVisuals}
    \vspace{-10pt}
\end{figure}

\begin{table*}
\vspace{1em}
    \centering
    \footnotesize
    \begin{tabularx}{\textwidth}{l | l || X | X | X | X | X | X || X | X | X | X | X || X }
        & & \multicolumn{5}{c}{Planning Time (ms)} & \multicolumn{5}{c}{Solution Cost (arclength)} & \\
        & System & Mean & Q1 & Med. & Q3 & 95\% & 100\%
        & Mean & Q1 & Med. & Q3 & 95\% & Succ.\\
        \hline
        \hline
            \parbox[t]{0.3mm}{\multirow{6}{*}{\rotatebox[origin=c]{90}{Panda}}}
     & Curobo & 62.63 & 25.92 & 30.08 & 34.38 & 281.63 & 1537.75 & 7.52 & \cellcolor{gray!10}\textbf{5.45} & 6.89 & 8.96 & 14.09 & \cellcolor{gray!10}\textbf{100.00\%} \\
     & OMPL-RRTC & 2.54 & 1.11 & 1.13 & 1.17 & 6.42 & 257.99 & 9.87 & 6.36 & 8.98 & 11.87 & 18.48 & \cellcolor{gray!10}\textbf{100.00\%} \\
     & GTMP Akima & 0.34 & 0.29 & 0.34 & 0.52 & --- & --- & 7.71 & 6.41 & 8.35 & 10.90 & --- & 79.11\% \\
     & GTMP Straight & 0.35 & 0.30 & 0.33 & 0.43 & --- & --- & 10.29 & 9.22 & 10.07 & 12.12 & --- & 89.41\% \\
     & VAMP-RRTC & \cellcolor{gray!10}\textbf{0.17} & \cellcolor{gray!10}\textbf{0.03} & \cellcolor{gray!10}\textbf{0.06} & \cellcolor{gray!10}\textbf{0.13} & 0.81 & 3.08 & 8.05 & 5.80 & 7.04 & 9.46 & 14.84 & \cellcolor{gray!10}\textbf{100.00\%} \\
     & pRRTC & 0.49 & 0.42 & 0.48 & 0.53 & \cellcolor{gray!10}\textbf{0.65} & \cellcolor{gray!10}\textbf{2.32} & \cellcolor{gray!10}\textbf{6.46} & 5.50 & \cellcolor{gray!10}\textbf{6.23} & \cellcolor{gray!10}\textbf{7.17} & \cellcolor{gray!10}\textbf{9.17} & \cellcolor{gray!10}\textbf{100.00\%} \\
    \hline
    \parbox[t]{0.3mm}{\multirow{2}{*}{\rotatebox[origin=c]{90}{Fetch}}}
     & VAMP-RRTC & 11.93 & \cellcolor{gray!10}\textbf{0.52} & 1.90 & 8.52 & 56.48 & 368.69 & 21.34 & 16.20 & 20.49 & 25.19 & 34.54 & \cellcolor{gray!10}\textbf{100.00\%} \\
     & pRRTC & \cellcolor{gray!10}\textbf{2.32} & 0.85 & \cellcolor{gray!10}\textbf{1.15} & \cellcolor{gray!10}\textbf{2.18} & \cellcolor{gray!10}\textbf{6.90} & \cellcolor{gray!10}\textbf{56.67} & \cellcolor{gray!10}\textbf{15.26} & \cellcolor{gray!10}\textbf{11.28} & \cellcolor{gray!10}\textbf{13.88} & \cellcolor{gray!10}\textbf{18.00} & \cellcolor{gray!10}\textbf{25.82} & \cellcolor{gray!10}\textbf{100.00\%} \\
    \hline
    \parbox[t]{0.3mm}{\multirow{2}{*}{\rotatebox[origin=c]{90}{Baxter}}}
     & VAMP-RRTC & 18.53 & 4.47 & 8.67 & 18.25 & 66.53 & 398.39 & 19.47 & 14.44 & 18.05 & 23.32 & 32.75 & \cellcolor{gray!10}\textbf{100.00\%} \\
     & pRRTC & \cellcolor{gray!10}\textbf{4.82} & \cellcolor{gray!10}\textbf{2.44} & \cellcolor{gray!10}\textbf{3.23} & \cellcolor{gray!10}\textbf{4.59} & \cellcolor{gray!10}\textbf{12.49} & \cellcolor{gray!10}\textbf{95.46} & \cellcolor{gray!10}\textbf{13.12} & \cellcolor{gray!10}\textbf{10.00} & \cellcolor{gray!10}\textbf{12.12} & \cellcolor{gray!10}\textbf{15.22} & \cellcolor{gray!10}\textbf{21.57} & \cellcolor{gray!10}\textbf{100.00\%} \\
    \hline
    \end{tabularx}
    \caption{
        Results for all robots and planners. Planning times shown in \textbf{milliseconds}, cost shown in units of configuration space.
    }
    \label{tbl:results}
\end{table*}

\begin{figure*}[t!] %
    \centering
    \begin{subfigure}[b]{0.49\textwidth}
        \centering
        \includegraphics[width=1\linewidth]{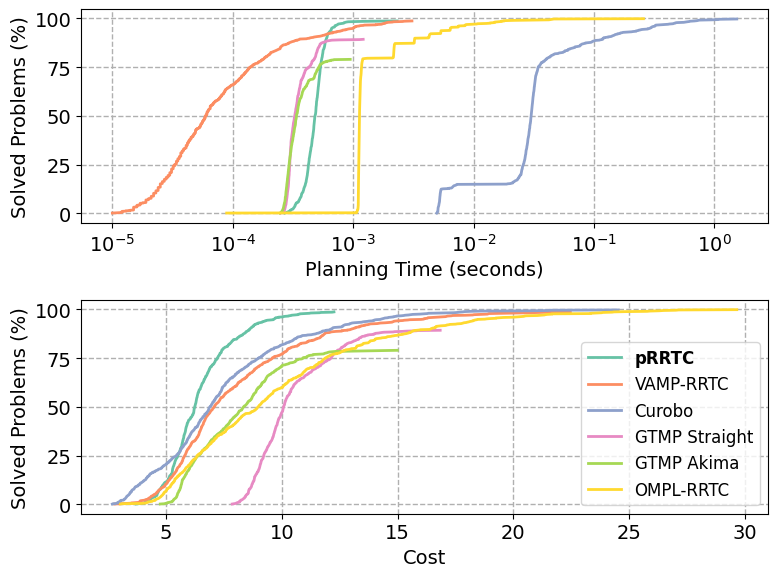}
        \vspace{-2em}
        \caption{MotionBenchMaker planning time and cost on 7 DoF Panda. 
        All times are shown on a \textbf{logarithmic} scale. pRRTC achieves lower cost initial solutions while performing on the order or better than state of the art on the most difficult problems.
        pRRTC is the first planner to solve all problems, and only pRRTC and VAMP-RRTC achieve both sub-millisecond planning time and 100\% solve rate.}
        \label{fig:panda}
    \end{subfigure}
    \hfill %
    \begin{subfigure}[b]{0.49\textwidth}
        \centering
        \includegraphics[width=1\linewidth]{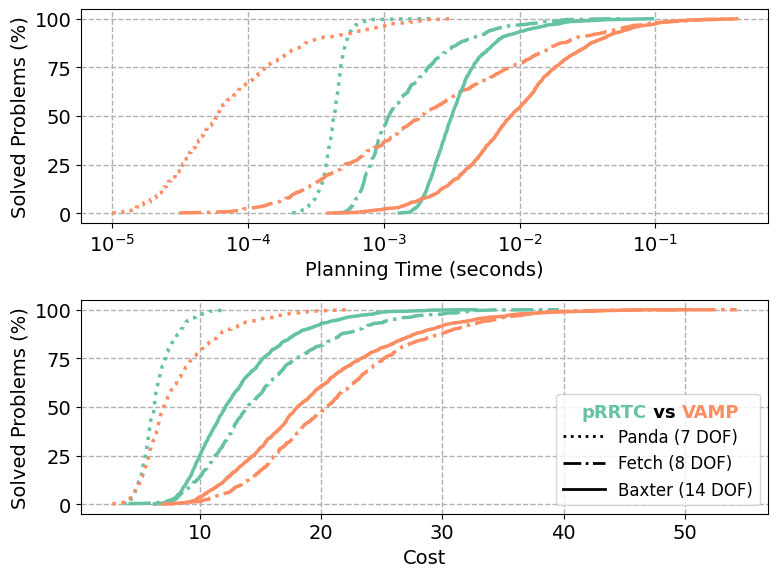}
        \vspace{-2em}
        \caption{
            MotionBenchMaker planning time and cost across all robots: the 7 DoF Panda, 8 DoF Fetch and 14 DoF Baxter.
            All times are shown on a \textbf{logarithmic} scale. As DoF and problem complexity grows, pRRTC performs better relative to VAMP-RRTC due to the speedups achieved via parallelism both in terms of average planning time as well as average final path cost.
        }
        \label{fig:allprobs}
    \end{subfigure}
    \caption{Empirical Cumulative Distribution Functions (ECDFs) for all Software Benchmarks.}
    \label{fig:cdfs}
    \vspace{-10pt}
\end{figure*}

\begin{figure}[t]
\vspace{1em}
    \centering
    \includegraphics[width=1\linewidth]{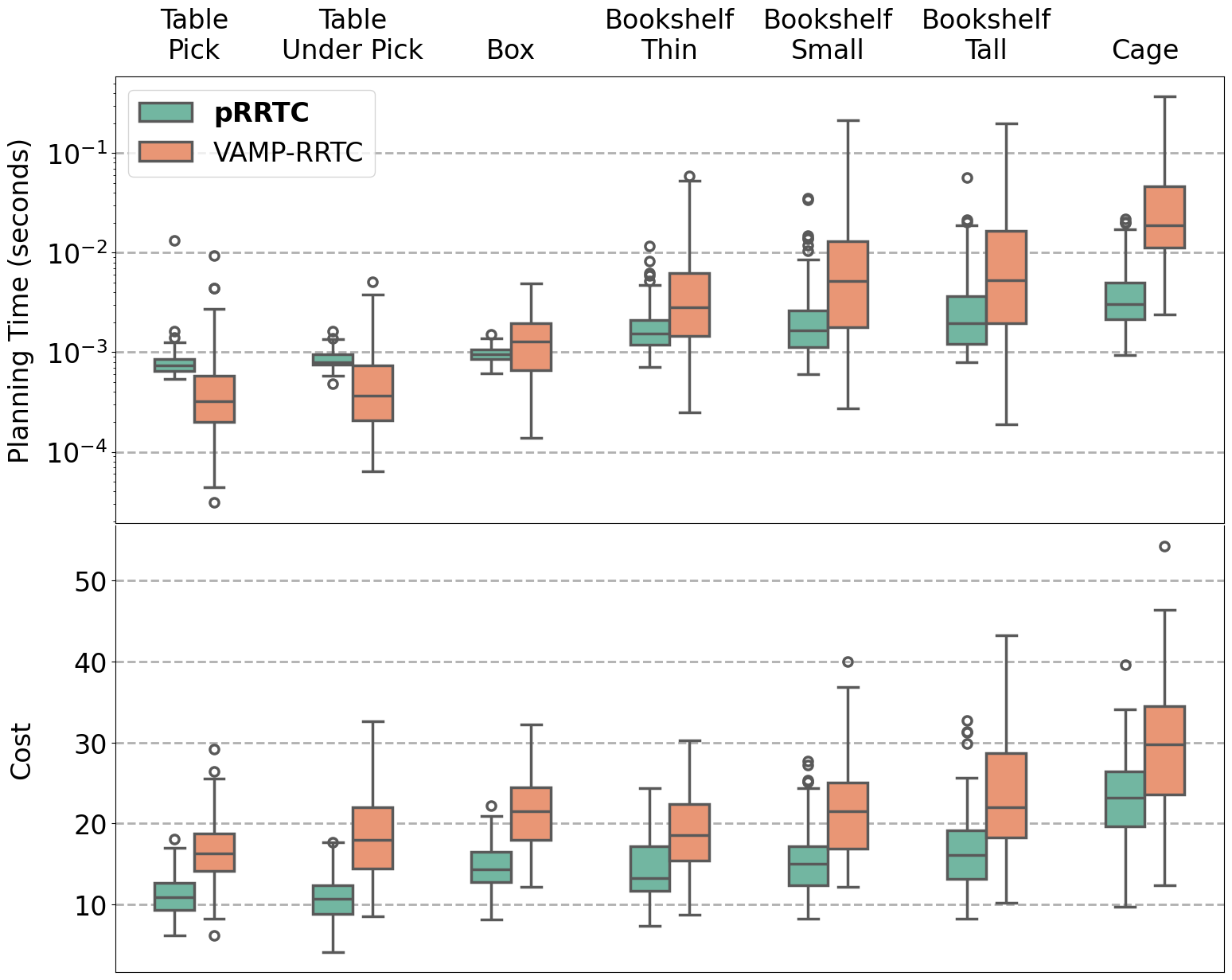}
    \caption{
        MotionBenchMaker planning time and cost by problem on Fetch (8 DoF). Problems are ordered from left to right by increasing complexity. 
        All times are shown on a \textbf{logarithmic} scale. pRRTC achieves a 5× speedup in average planning time, 10× speedup on the computationally challenging \emph{Cage} problems, and 8.4× decrease in planning time standard deviation.
    }
    \label{fig:fetch}
\end{figure}

We evaluated the planners on a set of realistic, challenging problems provided by the MotionBenchMaker dataset~\cite{chamzas2021motionbenchmaker}. The robots used were the 7 DoF Franka Emika Panda, 8 DoF Fetch, and 14 DoF Rethink Robotics Baxter. We compare all five planners on Panda. We also compare pRRTC and VAMP-RRTC on Fetch and Baxter. Curobo and GTMP are omitted from the Fetch and Baxter experiments as they do not support these robots. OMPL-RRTC is also omitted as it has been shown to be orders of magnitude slower than VAMP-RRTC on MotionBenchMaker~\cite{thomason2024motions} and on our initial Franka evaluation. For Panda and Fetch, MotionBenchMaker incorporates a diverse set of seven environments each with 100 problems which test the planner's ability to operate on table surfaces, reach into varying positions of bookshelves, and reach in highly constrained environments. For Baxter the dataset includes three bookshelf reaching scenes each with 500 problems. These benchmarks align with those used in~\cite{thomason2024motions,le2024global}. Visualizations of the environments are shown in Figure~\ref{fig:motionBenchMakerVisuals}.
Videos of our experiments are available online\footnote{\url{https://youtu.be/okpqftLB6P8}}.

\subsection{Software Benchmarks}

A summary of all results is shown in~\cref{tbl:results}. We present success rate versus performance curves of planning time and solution cost across all environments for the 7 DoF Panda robot using all baselines in~\cref{fig:panda}. We also present the comparison of pRRTC vs. VAMP-RRTC across all environments and robots in~\cref{fig:allprobs}.

On the 7 DoF Panda problems (\cref{fig:panda}), pRRTC is on average 128× faster than Curobo, 5× faster than OMPL-RRTC, 3× slower than VAMP-RRTC, and 1.4× slower than GTMP.  Compared to GTMP, pRRTC achieves a 100\% solve rate, making pRRTC and VAMP the only planners that solved all problems while averaging sub-millisecond planning time. In addition, pRRTC has the best worst case performance, that is, it solves all problems in the fastest amount of time, and produces the lowest average cost paths.

As we scale to larger DoF problems (\cref{fig:allprobs}), pRRTC outperforms VAMP-RRTC on average planning time, consistency, and initial path cost. On average, for the 8 DoF Fetch, pRRTC offers 5× speedup compared to VAMP-RRTC. We also present the distribution of results per-environment in \cref{fig:fetch} and observe that the benefit of pRRTC grows in more constrained environments. For example, on the \emph{Cage} problem pRRTC provides a 10× improvement on average planning time. pRRTC also has an 8.4× decrease in the standard deviation of planning time compared to VAMP-RRTC across all environments. This demonstrates pRRTC is not only a faster planner on average, but it also offers much more reliable performance. The advantage of pRRTC's parallel design also extends to path cost. On average, pRRTC produces initial paths with a 1.4× decrease in cost compared to VAMP-RRTC while showing a 1.3× decrease in the standard deviation of cost. Thus, pRRTC consistently finds lower cost initial paths.

Finally, for the 14 DoF Baxter, we see a similar trend of performance with pRRTC providing a 3.9× average planning time speedup over VAMP-RRTC and achieving a faster planning time on 93\% of problems.

\subsection{Ablation Studies}
\begin{figure}[!t]
\vspace{1em}
    \centering
    \includegraphics[width=1\linewidth]{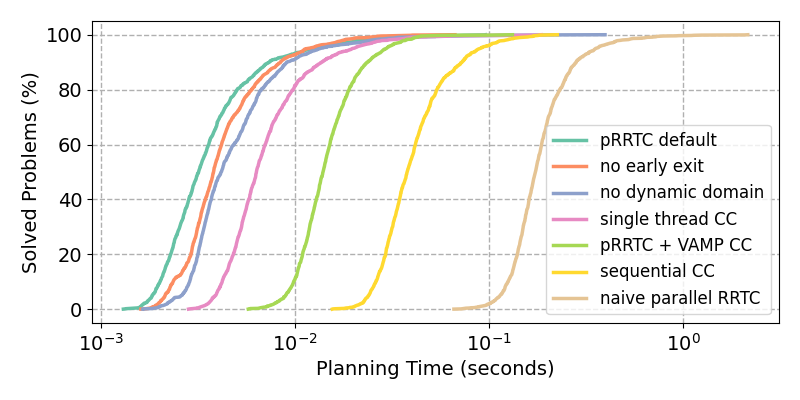}
    \caption{
        MotionBenchMaker planning time across different ablations of pRRTC on Baxter (14 DoF). All times are shown on a \textbf{logarithmic} scale. Each one of the design choices individually improved the efficiency of pRRTC, with their combined effects achieving a 41× speedup over a naive parallel baseline. 
    }
    \label{fig:ablation}
    \vspace{-10pt}
\end{figure}

\cref{fig:ablation} uses the 14 DoF Baxter for an ablation analysis of key design choices and optimizations we implement in pRRTC.
We find that our design choices all individually contribute to our overall speedup in planning time. The most significant gain came from checking discretized motions in parallel rather than sequentially, which resulted in an average 9.4× speedup. Replacing CPU-based FK and collision checking routines (e.g., those used in VAMP~\cite{thomason2024motions}) with customized GPU-optimized kernels for pRRTC provided a 3.3× average speedup. Using four threads per configuration to perform FK and CC, rather than a single thread, improved performance by an average of 1.8×. Incorporating dynamic domain sampling~\cite{yershova2005dynamic} contributed an average 1.3× gain. Enabling early termination of the CC routine upon detecting a collision led to an average 1.1× speedup. Finally, the combined impact of these optimizations is an average speedup of 41× over a naive baseline, showing how important careful co-design is for GPU-accelerated performance. 

\begin{figure}[!t]
    \centering
    \includegraphics[width=1\linewidth]{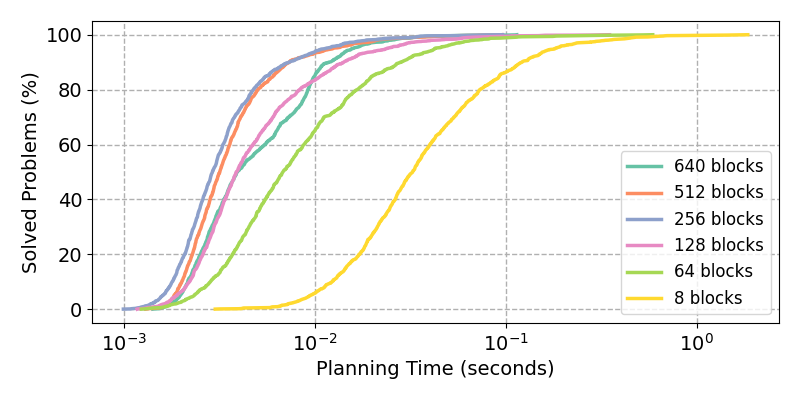}
    \caption{MotionBenchMaker planning time across varying block level parallelism of pRRTC on Baxter (14 DoF). All times are shown on a \textbf{logarithmic} scale. The planner benefits from increased parallelism until saturation around 256 blocks, after which block parallelism overhead led to suboptimal planning time.}
    \label{fig:ablation_blocks}
\end{figure}

In addition to exploiting parallelism for FK and CC subroutines, the general block level parallelism of concurrent sampling and expansion also contributes to the efficiency of pRRTC. As shown in~\cref{fig:ablation_blocks}, we see the planning time decreases as the number of blocks rise from 8 to 256, with a 14× speedup attributed to using 256 concurrent blocks rather than 8. As the block count increase beyond 256, the system becomes saturated, with the overhead of managing block saturation leading to worse planning time. This again shows that careful hardware-software co-design is needed for maximal algorithmic performance.


\subsection{Hardware Validation}

In order to test the feasibility of pRRTC in real-time planning scenarios, we deployed it on dual 7 DoF Franka Panda arms, resulting in a total of 14 DoF. Our setup consists of the arms mounted on a platform, with an overhead Intel Realsense camera streaming depth images at 90 fps. We use Nvblox~\cite{millane2024nvblox} to continuously integrate depth images into a Euclidean Signed Distance Field (ESDF). For real-time planning, we adapt our collision checking to query the ESDF rather than check against geometric primitives.
To test real-time replanning we have the robots sweep back and forth while obstructing their paths with dynamic obstacles. In our scenario, pRRTC is able to achieve real-time replanning, which, when paired with the perception routines, is able to achieve an end-to-end pipeline frequency of 7.7 Hz. The complete hardware demonstration can be viewed in the supplemental video submission.

\section{Conclusion and Future Work} \label{sec:conclusion}
In this work, we present pRRTC, a GPU-based parallel RRT-Connect algorithm coupled with a SIMT-optimized parallel collision checker. pRRTC utilizes both low- and medium-level GPU parallelism, parallelizing both low-level primitives (collision checking and nearest neighbors) and expansion of the tree while introducing almost no additional computational or communication overhead. Compared to both CPU- and GPU-based state-of-the-art motion planners on the MotionBenchMaker dataset~\cite{chamzas2021motionbenchmaker} using robots with 7, 8, and 14 degrees of freedom, pRRTC provides as much as a 10× speedup and 5.4× reduction in planning time standard deviation while achieving a 1.4× reduction in average initial path cost compared to existing approaches. We deploy pRRTC onto dual 7 DoF Franka Panda arms and demonstrate real-time, collision-free motion planning with dynamic obstacles.

Our future work includes transforming pRRTC into an almost-surely asymptotically optimal sampling-based planner~\cite{gammell2021asymptotically}, as the parallelization of framework iteration, collision checking, and nearest neighbor search translates naturally into the path optimization setting, particularly for planners that benefit from fast edge validation~\cite{wilson2024nearest}. We are also interested in leveraging real-time re-compilation to enable more dynamic co-designed optimizations~\cite{hu2025cprrtc} and aim to further improve our full end-to-end pipeline for additional real-world demonstrations in future work.

\printbibliography{}

\end{document}